\documentclass{sigchi}

\usepackage{balance}       
\usepackage{graphics}      
\usepackage[T1]{fontenc}   
\usepackage{txfonts}
\usepackage{mathptmx}
\usepackage[breaklinks=true,letterpaper=true,colorlinks,bookmarks=false]{hyperref}
\usepackage{color}
\usepackage{booktabs}
\usepackage{textcomp}
\usepackage{bm}

\usepackage{microtype}        
\usepackage{ccicons}          

\usepackage{todonotes}

\def\plaintitle{FLAVA: Find, Localize, Adjust and Verify to Annotate LiDAR-Based Point Clouds}
\def\plainauthor{Tai Wang, Conghui He, Zhe Wang, Jianping Shi, Dahua Lin}

\def\plainkeywords{Point Cloud Annotation, LiDAR, Scene Understanding, Autonomous Driving}

\makeatletter
\def\url@leostyle{%
  \@ifundefined{selectfont}{
    \def\UrlFont{\sf}
  }{
    \def\UrlFont{\small\bf\ttfamily}
  }}
\makeatother
\def\pprw{8.5in}
\def\pprh{11in}

\definecolor{linkColor}{RGB}{6,125,233}
\hypersetup{%
  pdftitle={\plaintitle},
  pdfauthor={\plainauthor},
  pdfkeywords={\plainkeywords},
  pdfdisplaydoctitle=true, 
  bookmarksnumbered,
  pdfstartview={FitH},
  colorlinks,
  citecolor=black,
  filecolor=black,
  linkcolor=black,
  urlcolor=linkColor,
  breaklinks=true,
  hypertexnames=false
}

\makeatletter
\def\@copyrightspace{\relax}
\makeatother

\begin{document}

\title{\plaintitle}

\numberofauthors{1}
\author{%
  \alignauthor{Tai Wang$^{1, 2}$~~~~ Conghui He$^2$~~~~ Zhe Wang$^2$~~~~ Jianping Shi$^2$~~~~ Dahua Lin$^{1, 2}$\\
    \affaddr{$^1$The Chinese University of Hong Kong, $^2$Sensetime}\\
    \email{\{wt019, dhlin\}@ie.cuhk.edu.hk}, \{heconghui, wangzhe, shijianping\}@sensetime.com}\\
}

\maketitle


\begin{abstract}
Recent years have witnessed the rapid progress of perception algorithms on top of LiDAR, a widely adopted sensor for autonomous driving systems. These LiDAR-based solutions are typically data hungry, requiring a large amount of data to be labeled for training and evaluation. However, annotating this kind of data is very challenging due to the sparsity and irregularity of point clouds and more complex interaction involved in this procedure. To tackle this problem, we propose FLAVA, a systematic approach to minimizing human interaction in the annotation process. Specifically, we divide the annotation pipeline into four parts: find, localize, adjust and verify. In addition, we carefully design the UI for different stages of the annotation procedure, thus keeping the annotators to focus on the aspects that are most important to each stage. Furthermore, our system also greatly reduces the amount of interaction by introducing a light-weight yet effective mechanism to propagate the annotation results. Experimental results show that our method can remarkably accelerate the procedure and improve the annotation quality.
\end{abstract}


\printccsdesc


\section{Introduction}
\label{sec:introduction}
LiDAR is widely used in today's autonomous driving systems. It can provide accurate spatial information of the 3D environment, and thus assist the scene understanding and decision-making process of the system. 
In recent years, a lot of perception algorithms using deep learning have emerged to handle this kind of data~\cite{VoxelNet,PointPillars,PointRCNN,tracking_baseline,ssn,reconfig_voxels}, which are significantly superior to monocular and stereo approaches in application. 
The rapid progress of these algorithms is supported by several challenging benchmarks built on multiple open datasets~\cite{KITTI,nuScenes,Lyft,waymo}. 
However, although a decent amount of data has been released, the actual product deployment still needs more data with accurate labels to feed the algorithms. The only publicly accessible tools for annotation like~\cite{3DBAT} are still very coarse, especially in terms of annotation accuracy, which limits the research progress in this field.

\begin{figure}
   \centering
   \includegraphics[width=1.0\linewidth]{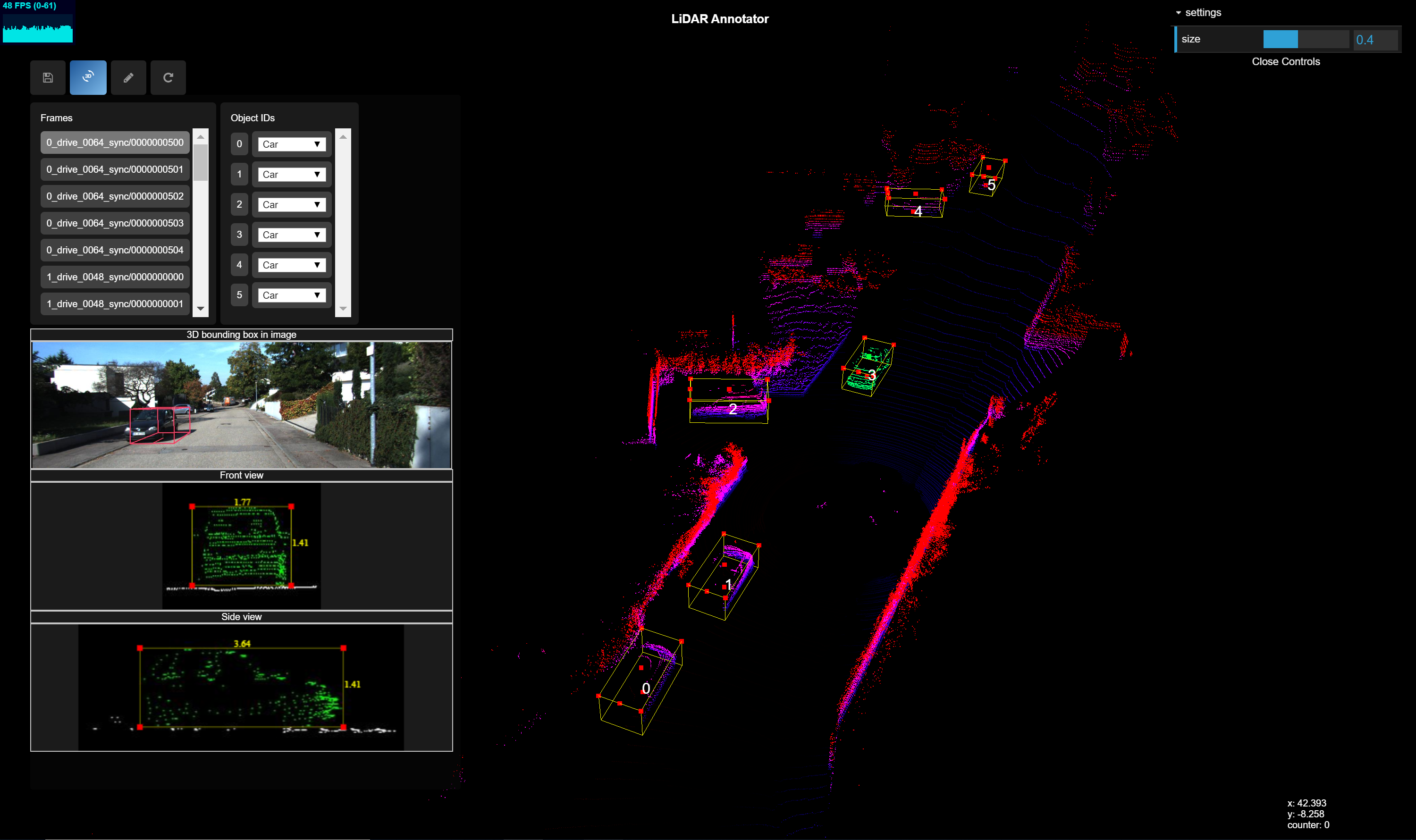}
   \caption{A screenshot of our annotation tool. Best viewed in color.}
   \label{fig: teaser}
\end{figure}

\begin{figure*}
   \centering
   \includegraphics[width=1.0\linewidth]{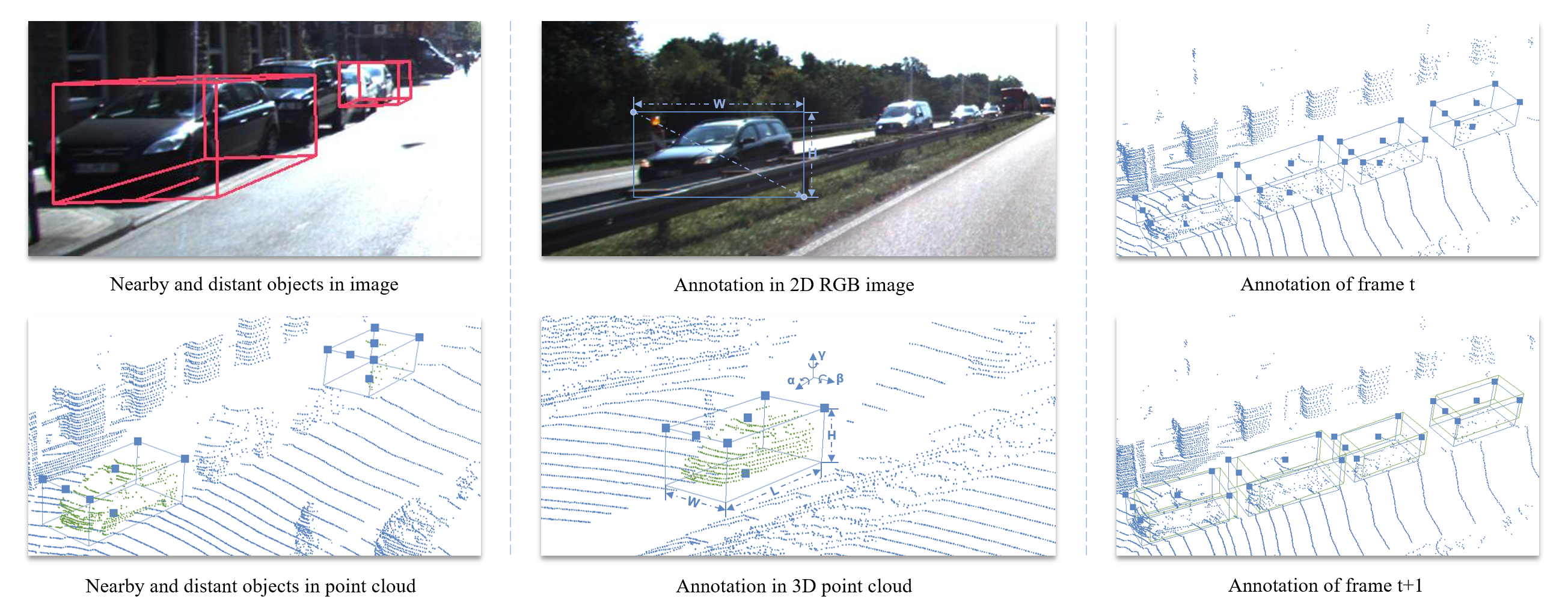}
   \caption{Challenges of annotating LiDAR-based point clouds. From left to right: Sparse and irregular spatial distribution of point cloud poses great challenges to annotation; Operations are much more complex when annotating 3D point cloud; The input data and annotated labels of consecutive frames both have strong sequential correlation.}
   \label{fig: challenges}
\end{figure*}

While there are many existing approaches to efficiently annotating RGB images~\cite{VIA,PolygonRNN,PolygonRNN++,Curve-GCN}, not much work has focused on 3D annotation tasks due to their more complex cases (Figure~\ref{fig: challenges}).  
First of all, it is difficult to identify all the objects of interest correctly in the sparsely and irregularly distributed point cloud. Further, the operation complexity is relatively high considering the larger degree of freedom (DoF) in the procedure, such as the need for annotating height and steering angle of objects, thus requiring customized UI design to achieve accurate annotation. Finally, there exists sequential correlation between consecutive frames, which can be leveraged to reduce the operations of annotators.
A few recent works~\cite{LATTE,pretrained_generation,Autolabeling} noticed these problems, but they mainly used some algorithm-assisted semi-automatic ways to improve the efficiency of annotation rather than focused on the human-computer interaction in this process. Actually, these algorithms are not much efficient and convenient in practical use considering the equipment provided to annotators. Most of them need GPUs to train models and are not able to run smoothly on an ordinary laptop.

\begin{figure*}
   \centering
   \includegraphics[width=1.0\linewidth]{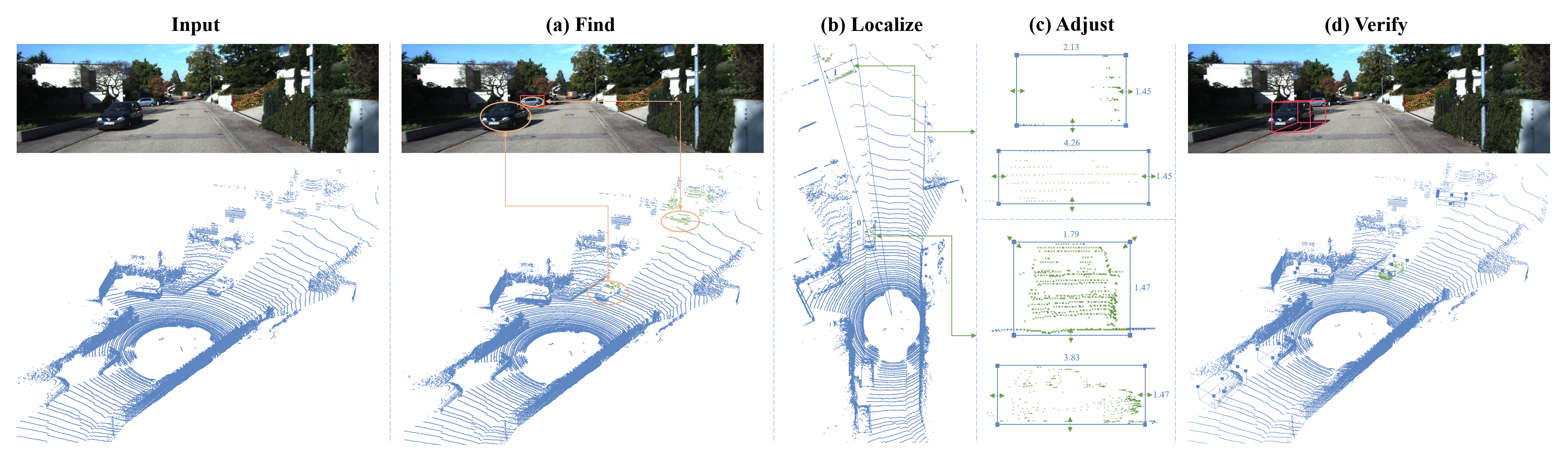}
   \caption{An overview of our pipeline. Our proposed FLAVA is a systematic approach to minimizing human interaction when annotating LiDAR-based point clouds. It can be divided into four steps: find, localize, adjust and verify. Given the input RGB image and point cloud, we first find and localize the object of interest primarily in a top-down way, then adjust the bounding box in the projected views of local point cloud, and finally verify the annotation in the RGB image and the perspective view of point cloud. Note that in this process, the annotation task is ultimately implemented on the point cloud data, and the semantic information of the RGB image can effectively assist annotators in the preliminary localizing and final verification.}
   \label{fig: pipeline}
\end{figure*}

In this work, we target on the human-computer interaction in the process of 3D annotation, especially the annotation used for detection and tracking. We aim at tackling two difficulties in this process from the perspective of annotators: the difficulty of identifying objects correctly in the global scene at the beginning and the difficulty of accurately labeling objects after primarily localizing them. Specifically, we propose FLAVA, a systematic annotation pipeline to minimize the annotator's operations, which can be divided into four steps: find, localize, adjust and verify. As shown in Figure~\ref{fig: pipeline}, to label a 3D bounding box, we \emph{find} the targets in a top-down way at first and then \emph{localize} it primarily in the top view, where the first difficulty is needed to be tackled. Subsequently, after the height is automatically computed, we \emph{adjust} the box on the projected view of the local point cloud in a bottom-up way to solve the second problem. Finally, the semantic information of the RGB image and the perspective view of the point cloud can be combined to \emph{verify} the results.

Apart from the whole constructive pipeline, we also design a UI tailored to these four stages (Figure~\ref{fig: teaser} and \ref{fig: UI}). The UI has several appealing functions, such as various zoomable views for multimodal data, highlight of local point clouds and length specification, which keep the annotators focusing on the most important tasks at each stage and thus ensure the accuracy of annotated results. Furthermore, we introduce a mechanism to propagate the annotated results between objects and consecutive frames. With this mechanism, most 3D annotation cases can be basically simplified as concise operations of 2D boxes in the top view,  which significantly reduces unnecessary repeated operations.

We evaluated the proposed annotation method with several sequences collected from KITTI raw data. Compared with our baseline, it can not only accelerate the annotation speed by 2.5 times, but also further improve the quality of the labels, as measured by 27.50\% high 3D average precision, and 9.88\% high bounding box IoU.

Our contributions of this work are summarized as follows:
\begin{itemize}
\item We start from the human habit of understanding a scene, and propose a systematic annotation pipeline, namely FLAVA, to tackle the two key problems in 3D annotation tasks, identifying objects correctly and annotating them accurately.
\item We designed a clear UI and annotation transfer mechanism according to the characteristics of data and tasks, which makes it more convenient for annotators to concentrate on much simpler work at each stage and accomplish it with fewer operations.
\item We tested the proposed annotation method on the KITTI dataset, and proved its remarkable effect on the efficiency and quality of labeling. Detailed ablation studies reveal the significance of different functions on this issue.
\end{itemize}

\section{Related Work}
\label{sec:related}
\subsection{LiDAR-based benchmarks}
In recent years, LiDAR has been widely used in various autonomous driving systems. In order to promote the development of this field, many open datasets have been released. Several benchmarks of various tasks are set up on top of them, including 3D object detection, 3D object tracking and point cloud semantic segmentation. One of the pioneers in this aspect is the KITTI dataset~\cite{KITTI}, which has about 15000 frames of data in 22 scenes for training and testing, including about 200K 3D boxes. Afterwards, two large-scale datasets named nuScenes~\cite{nuScenes} and Lyft~\cite{Lyft} were released. They shared similar data formats and provided point cloud data of consecutive frames. NuScenes dataset is split in 700/150/150 scenes for training/validation/testing respectively. There are overall 1.4M annotated 3D boxes, far more than KITTI’s 200K boxes. Lyft dataset has 180 and 218 scenes for training and testing. Both of them have more data, more object categories and richer scenes. Recently, Waymo open dataset~\cite{waymo} has been released and it is currently the largest dataset along them. In addition, it is worth noting that Waymo uses a mid-range lidar and four short-range lidars, which are different from the 64-line velodyne used by KITTI and the 32-line velodyne used by nuScenes and Lyft.

On the basis of these open datasets, many algorithms have emerged to solve these 3D tasks, such as \cite{MV3D,VoxelNet,PointPillars,PointRCNN,ssn,reconfig_voxels} for 3D detections and \cite{tracking_baseline,prob_tracking} for 3D tracking, However, despite these open datasets, the actual product adoption still needs more data support to ensure the stability and security of algorithms. Moreover, when the configuration of the lidar changes, for example, the location is different or the number of lines is different, the model needs new data for training and tuning. All of these show that an efficient annotation method is still an important demand in this field.

\subsection{Annotation tools}
As data plays an increasingly important role in various fields of computer vision, assisted labeling has gained great popularity. For images and videos annotation, VIA~\cite{VIA} proposed a simple and standalone annotation tool for image, audio and video. Polygon-RNN~\cite{PolygonRNN,PolygonRNN++} trained a recurrent CNN to inference the polygonal mask on the image to assist the annotation for semantic segmentation. Curve-GCN~\cite{Curve-GCN} further improved the efficiency of generating polygon vertices and achieved real-time interaction. For semi-automatic annotation tailored to autonomous driving applications, BDD100K~\cite{BDD100K} proposed a set of tools to annotate bounding boxes and semantic masks on RGB images. It also leveraged pretrained detectors to accelerate the annotation for 2D detection. Few works focused on the annotation in LiDAR-based point clouds. \cite{pretrained_generation} presented a method to generate ground truths via selecting spatial seeds assisted by pretrained networks. \cite{active_learning} utilized active learning to train 3D detectors while minimizing human annotation efforts. \cite{Autolabeling} proposed to autolabel 3D objects from pretrained off-the-shelf 2D detectors and sparse LiDAR data. LATTE~\cite{LATTE} used sensor fusion, one-click annotation and tracking to assist point cloud annotation in the bird view. However, although there exist these works investigating how to accelerate this process, most of them tried to use algorithms to achieve it instead of diving into the details of 3D interactions. Furthermore, most of them are not much efficient and practical regarding the equipment deployed to the annotators.

\begin{table*}
	\begin{center}
	\begin{tabular}{c|c|c|c}
	\hline
	Method & Data & Task & Characteristics\\
	\hline\hline
	VIA~\cite{VIA} & Image, audio and video & Multi-task & Simple and standalone\\
	\hline
	Polygon-RNN~\cite{PolygonRNN,PolygonRNN++} & Image & Semantic segmentation & Recurrent CNN, polygonal mask\\
	\hline
	Curve-GCN~\cite{Curve-GCN} & Image & Semantic segmentation & GCN, predict vertices simultaneously\\
	\hline
	BDD 100K~\cite{BDD100K} & Image and video & Multi-task & 2D pretrained detectors, the largest dataset\\
	\hline
	GT Generation~\cite{pretrained_generation} & Point Cloud & 3D detection & 3D pretrained detectors\\
	\hline
	LiDAR Active Learning~\cite{active_learning}  & Point Cloud & 3D detection & Active learning\\
	\hline
	Autolabeling~\cite{Autolabeling} & Point Cloud & 3D detection & Signed distance fields (SDF)\\
	\hline
	LATTE~\cite{LATTE} & Point Cloud & BEV detection & Mark-RCNN, Clustering, Kalman filter\\
	\hline
	\end{tabular}
	\end{center}
           \vspace{1ex}
	\caption{A few published works for accelerating annotation procedure. Compared to them, our FLAVA focuses on the complex 3D interaction involved in the annotation of LiDAR-based point clouds. It serves as a systematic pipeline to provide accurate labels for 3D detection and tracking.}
	\label{tab: AP}
\end{table*}

\subsection{Data generation from LiDAR simulation}
Because the annotation of point clouds is challenging and time-consuming, many research efforts aim at building simulation environment to obtain enough data for training neural networks. \cite{LiDAR_generator} proposed a framework to produce point clouds with accurate point-level labels on top of a computer game named GTA-V. This kind of simulated data can be combined with the data from the real world to feed algorithms (\cite{SqueezeSeg,SqueezeSegv2,LiDAR_generator}). CARLA~\cite{CARLA} and AutonoVi-Sim~\cite{AutonoVi-Sim} also tried to simulate the LiDAR point cloud data from the virtual world. However, their primary target is to provide a platform for testing algorithms of learning and control for autonomous vehicles instead of augmenting specific LiDAR data. Furthermore, due to the difference of spatial distribution between the simulated and real data, the model trained with these platforms performs poorly on the real-world data. Although some researchers have made great progress in this domain adaptation problem, the gap was just reduced but not closed. Therefore, an efficient annotation pipeline used to collect data from the real world is still a critical need.
\section{Methodology}
\label{sec:methodology}
\textbf{Overview} Object detection and tracking in LiDAR-based point clouds are very important tasks for the 3D perception system of autonomous driving. Current algorithms need to be trained and tested with manually labeled data to accomplish these tasks. Specifically, in this type of annotation task, the annotator needs to correctly identify the object to be detected in the sparse point cloud first and then accurately label its position, size, orientation, category, and so on. Achieving both of them efficiently is not trivial due to the complex interaction involved in the procedure.
Our FLAVA is a systematic approach to addressing this issue. In this section, we will elaborate the four steps as well as the UI designs involved in the annotation pipeline (Figure~\ref{fig: pipeline} and \ref{fig: UI}), where the first two steps aim at identifying and localizing the objects primarily in a global view, the third step is to annotate accurately, and the final step is to ensure all the annotations are confident enough. Finally, we will present the annotation transfer mechanism used in our system that can greatly reduce unnecessary interactions.

\begin{figure}
   \centering
   \includegraphics[width=0.85\linewidth]{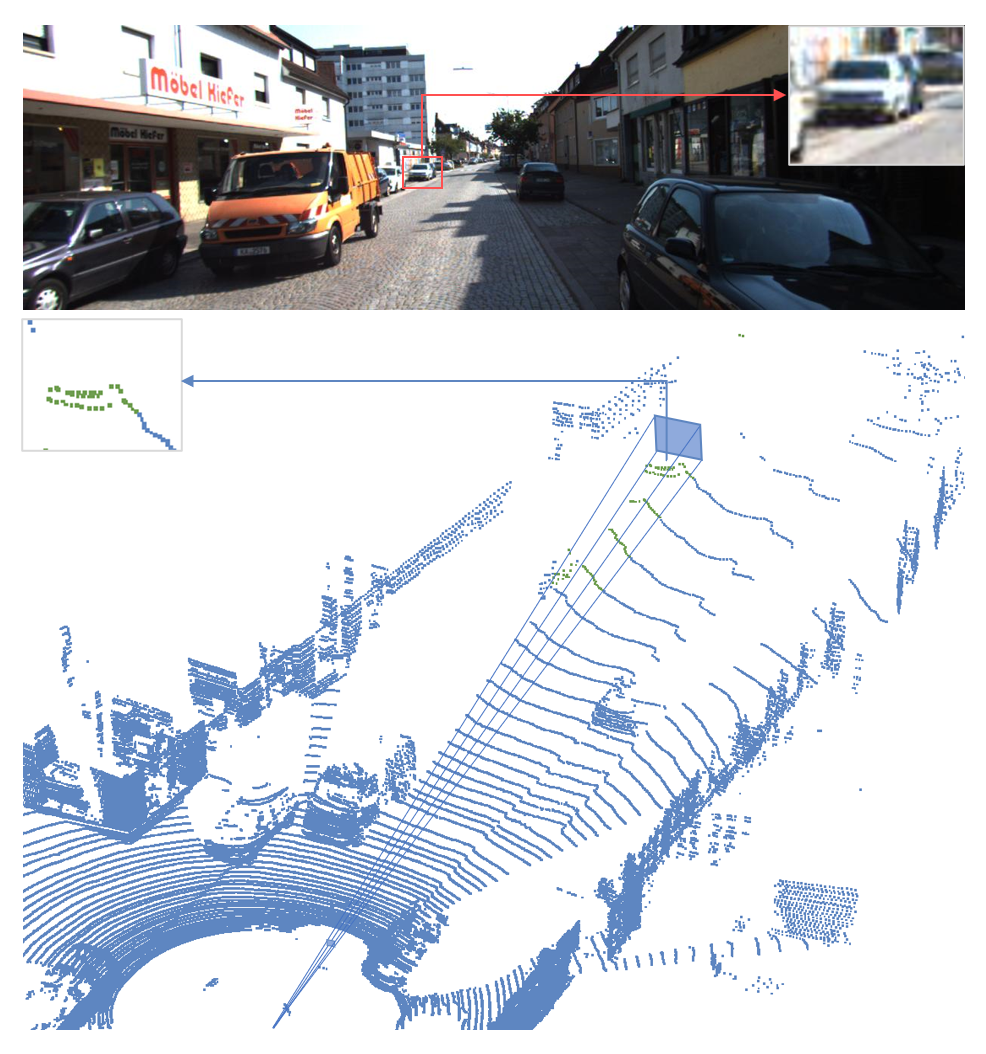}
   \caption{The frustum proposal is used to tackle the difficulty of finding and localizing distant objects. This example shows that some objects can be easily found in the RGB image, but it may be much more difficult to identify it in the point cloud directly.}
   \label{fig: frustum}
   \vspace{-2ex}
\end{figure}

\subsection{Find}
\begin{figure*}
   \centering
   \includegraphics[width=0.95\linewidth]{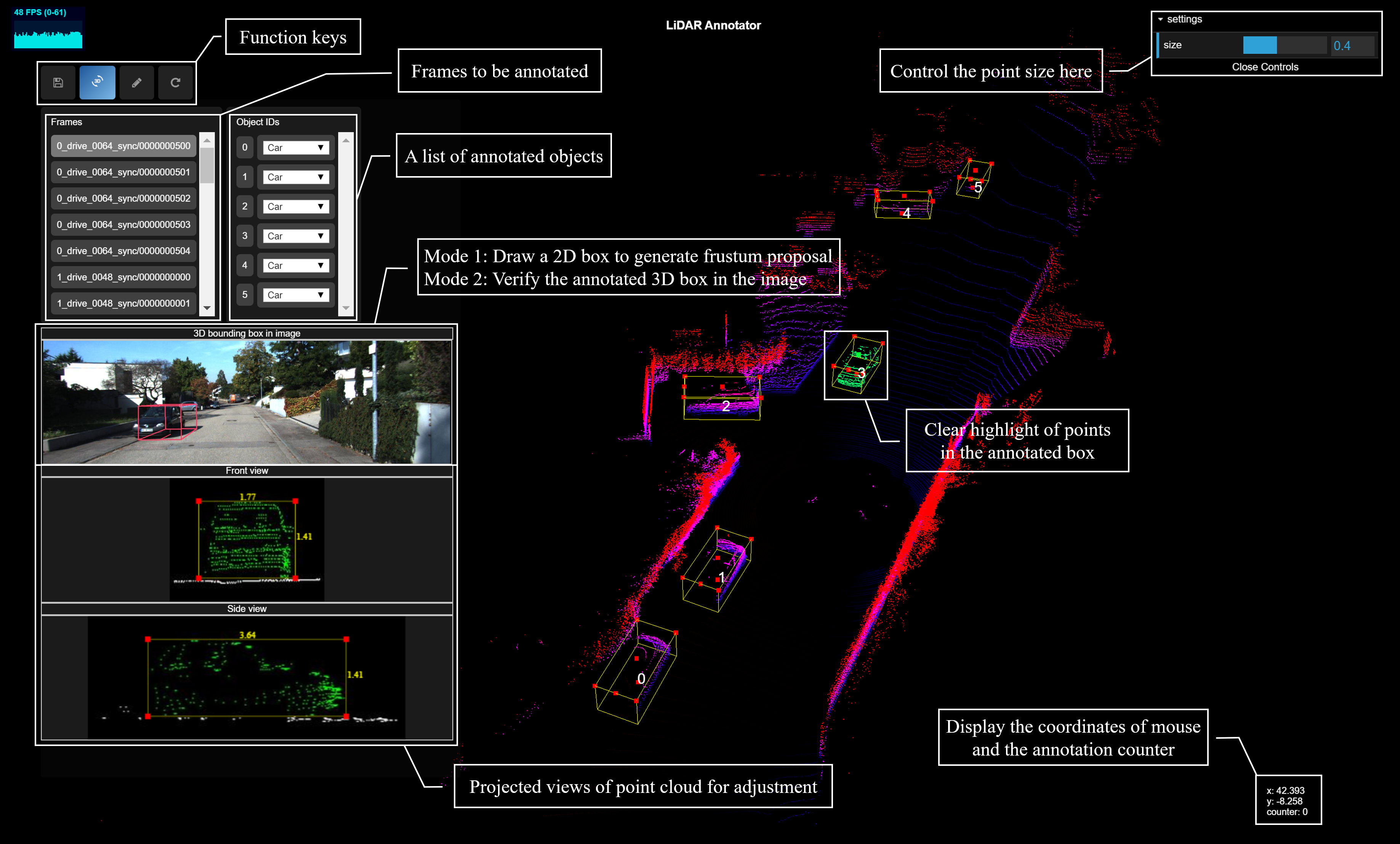}
   \caption{The labeling UI used by annotators to conduct our proposed FLAVA pipeline. The function keys from left to right are used to save annotations, switch to the 3D view, switch to the bird view and switch the RGB image mode between 1 and 2. Best viewed in color.}
   \label{fig: UI}
\end{figure*}

To begin with, we need to find target objects from the entire scene. Point clouds can accurately reflect the 3D physical environment in the real world; while RGB images can provide semantic information for human analysis. How to combine these two modes of data is a key problem. For the point cloud, apart from its perspective view, considering the particularity of the scenario, the objects we need to detect are basically on the ground, so the bird view of the point cloud is also a good starting global view for labeling, which can avoid the occlusion problem between objects that may exist in RGB images. 
For nearby and large objects, we can easily find them based on these two views of the point cloud, such as the object 3 in the Figure~\ref{fig: UI}.
For distant and small objects, it is difficult to identify them directly in the point cloud, and the semantic information of RGB images is needed. As the Figure~\ref{fig: frustum} shows, the object of interest may have only a few points obtained from LiDAR, but it can be found directly in the image. Therefore, we can leverage the corresponding frustum proposal~\footnote{The frustum proposal refers to the 3D search space lifted from a 2D bounding box in the image with near and far planes specified by depth sensor range.} in 3D space to find the object primarily.

\begin{figure}
   \centering
   \includegraphics[width=1.0\linewidth]{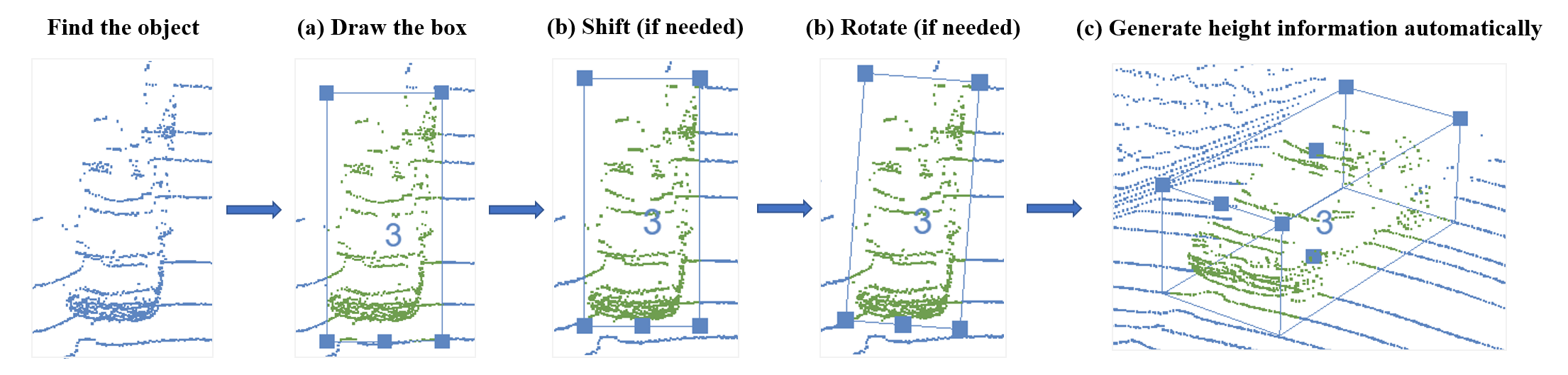}
   \caption{Localizing the object primarily can be conducted by three steps.}
   \label{fig: localize}
\end{figure}

Specifically, we first find the approximate position in the RGB image, and further identify which points are relevant by highlighting those within the generated frustum proposal and estimating its distance in the 3D environment (Figure~\ref{fig: pipeline}(a)(b) and Figure~\ref{fig: frustum}). In the process, we need to use the projection transformation from point cloud to image when constructing the frustum proposal:

\begin{equation}
{\rm y} = P_{rect}^{(i)}R_{rect}^{(0)}T_{velo}^{cam}{\rm x}
\label{eqn: projection}
\end{equation}

where ${\rm x} = (x,y,z,1)^T$ is a 3D point in the velodyne coordinate system, ${\rm y} = (u,v,1)^T$ is the projected coordinate in the camera image, $P_{rect}^{(i)}\in \mathbb{R}^{3\times 4}$ is the projection matrix after rectification corresponding to the i-th camera, $R_{rect}^{(0)}\in \mathbb{R}^{4\times 4}$ is the rectifying rotation matrix (expanded by appending a fourth zero-row and column, and setting $R_{rect}^{(0)}(4,4)=1$), $T_{velo}^{cam}$ is the rigid body transformation from velodyne coordinates to camera coordinates. After projected onto the image, the points falling into the 2D box in the RGB image will be highlighted for our reference, as shown in Figure~\ref{fig: pipeline}(a) and Figure~\ref{fig: frustum}. With explicitly marking the relevant points, we can basically identify which points belong to the object of our interest by combining the RGB image and the contextual information in the nearby region.

\subsection{Localize}
Once we have found the object of interest, what we need to do subsequently is to localize it. Primarily "finding" and "localizing" objects share similar characteristics in terms of visual perception. They both aim to correctly identify the targets from a global environment, and thus top-down methods should be more effective. Therefore, we still mainly focus on the bird view of the entire scene, supplemented by the perspective view. In terms of UI design, considering the large scope of the global scene and the importance of point cloud data, we also give it the largest area to display (Figure~\ref{fig: UI}). We divide the whole process into three parts: drawing bounding boxes in the bird view, adjusting their position and orientation, and finally generating height information automatically.

As is shown in the Figure~\ref{fig: localize}, we find the object of our interest at first, draw the bounding box in the top view, and then adjust its position and orientation by shifting and rotating without changing its size. As mentioned later, this will be the most core and simple operation throughout our annotation process, especially when the size and height of the box are initially determined. Note that after we draw the box, the front view and side view of the local point cloud will be updated. The orientation in the side view is very useful when determining whether we have annotated a correct orientation. Regarding the side view derived here is observed from the right side, the object facing right indicates our correct annotation (Figure~\ref{fig: pipeline}(c)).

\begin{figure*}
   \centering
   \includegraphics[width=1.0\linewidth]{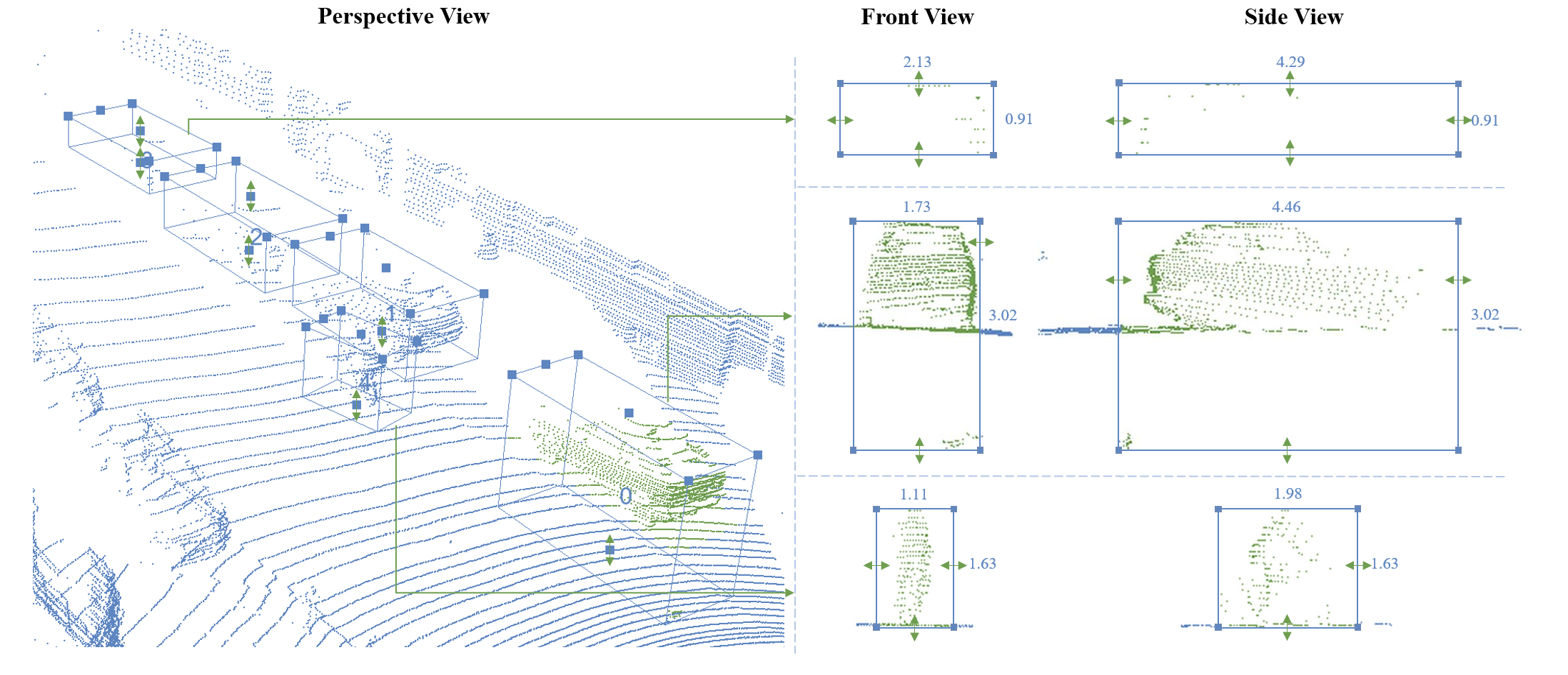}
   \caption{Comparison of adjusting boxes in the perspective view, front view and side view. The examples from top to bottom show the special cases that may be encountered: The lidar can only sweep the top part of the object and get sparse point cloud; There exist some noises influencing the automatic computation of height information; The points on the ground are usually involved in the automatically annotated 3D bounding box. In these cases, it is evident that adjusting the bounding box in the front view and side view is much more accurate and efficient than adjusting it in the perspective view.}
   \label{fig: adjust}
\end{figure*}

Finally, the height and 3D center of the box are automatically generated based on the highest and lowest points within the 2D box in the top view. The box we get here is an incompletely accurate one that tightly covers the point cloud vertically. For example, when a point cloud is swept only over the top half of an object, the position of the box we get may be skewed; when a point cloud is scanned more fully, the points on the ground or some noises may get involved (Figure~\ref{fig: adjust}). Therefore, in order to get a more accurate labeling result, we need to finetune the size and position of the box next.

\subsection{Adjust}
Unlike the previous two steps, when adjusting the box, the analysis of \emph{local} saliency is more important, which means that it would be better to be done in a bottom-up way. Here, we use the front view and side view of the local point cloud as the main data formats for our operations. As shown in Figure~\ref{fig: adjust}, this design is particularly important when distant objects need to be labeled. On the one hand, labeling distant objects are constrained by the 3D interactive environment, which makes it difficult to zoom in and observe them carefully; on the other hand, operating directly in the environment to annotate height information can also result in inadequate flexibility and accuracy. Instead, considering the incompleteness of the scanning and the symmetry of the object outline, the front view and side view of a local point cloud can best help the annotator to imagine the general shape of the object and pay attention to more details such as whether the points on the boundary are involved in, so that the annotator can draw a more accurate box. Note that by borrowing the idea of \emph{anchor} from detection algorithms\footnote{We usually set a reference box, namely anchor, for each category of objects to simplify the inference of box size in the detection algorithms.}, here we specify the length of each edge of the box in the projected views, which can make it convenient for annotators to compare their annotation with the reference box size and approximate the complete bounding box more reasonably.

\begin{figure}
   \centering
   \includegraphics[width=1.0\linewidth]{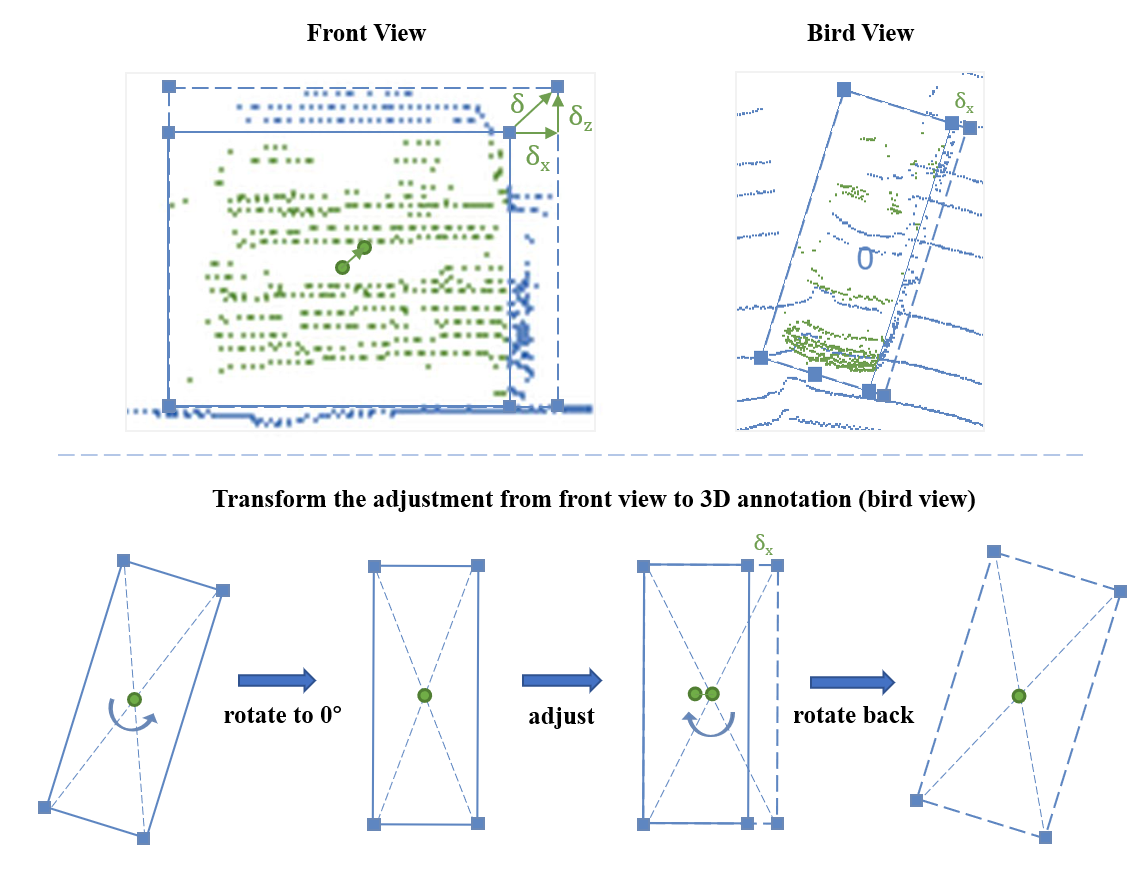}
   \caption{Illustration of transforming the adjustment in the front view to 3D annotation (in bird view). We split the adjustment into two orthogonal directions and transform them to our 3D annotation. The same method can be applied to the adjustment in the side view.}
   \label{fig: adjust_transform}
\end{figure}

To be more specific for the implementation, when finetuning boxes in the front view and side view of a local point cloud, we need to map the adjustment in the 2D view to the 3D box. Taking the case of front view as an example (Figure~\ref{fig: adjust_transform}), we split the adjustments in the 2D view into two orthogonal directions and transform the 3D box accordingly. For height adjustment, there is no particular coordinate transformation. For the operation in the horizontal direction, we first turn the box back to the 0° orientation, adjust its vertices coordinates, find a new center, and then rotate it back to the original orientation. Note that this example can be extended to any possible cases like resizing in other ways or shifting the box. Extension to the case in the side view is also straightforward, where we just need to simply apply the changes on the width to the length.

\subsection{Verify}
After adjusting, we need to verify the annotated box at the end. At this time, we can make full use of all kinds of modal data besides the projected views of the local point cloud for validation, including the various stereo perspectives of the 3D point cloud and RGB images. In this process, various zoomable views, highlight of local point clouds and length specification in the UI design are all important details to assist the annotator to verify (Figure~\ref{fig: UI}).

For the point cloud, we can switch to the perspective view for observation, especially when the point cloud is sparse, we need to further confirm whether the imaginary height of the box is reasonable in the global view. In addition, the projected view of the local point cloud can be used to further confirm whether the boundary and orientation of the labeled object are correct. For the RGB image, we use Eqn.~\ref{eqn: projection} to project eight vertices of the bounding box into the image, and verify the correctness of annotation with semantic information.

After the verification of various perspectives, if we need to adjust the position, orientation and size of the object, considering that the height adjustment in the third step has been very accurate, we specially fix the height information of the object (including the height of the box and its center). This detail will also be covered in the later part, in order to reduce unnecessary repeated operations in height adjustment and improve the stability of height annotation.

\begin{figure*}
   \centering
   \includegraphics[width=1.0\linewidth]{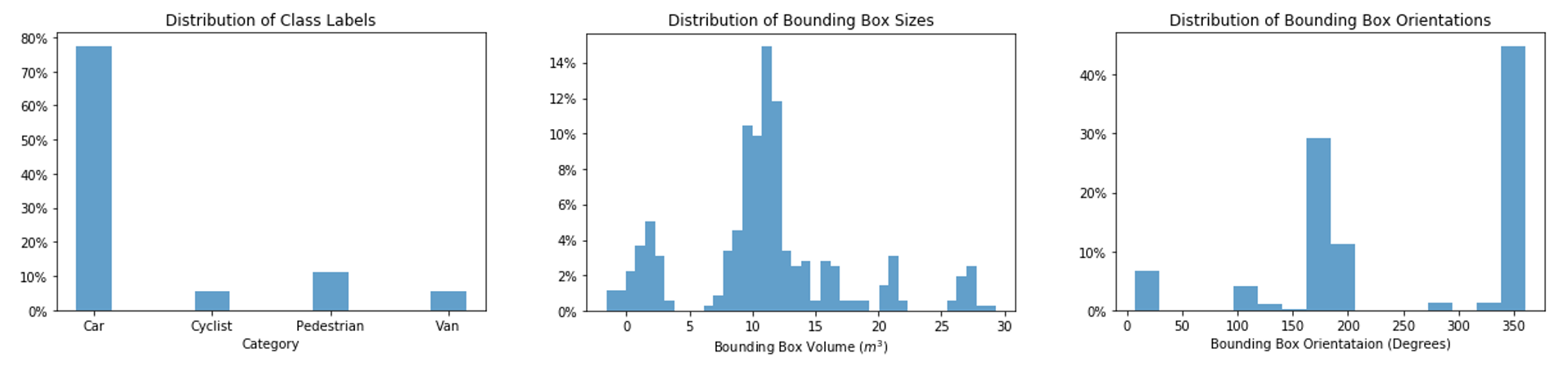}
   \caption{Distribution of class labels, bounding box sizes and bounding box orientations in our test benchmark.}
   \vspace{2ex}
   \label{fig: data_analysis}
\end{figure*}

\subsection{Annotation Transfer}
The previous four parts describe the labeling process for a single object or a single frame. In this section, we will describe the most important detail used throughout the labeling procedure, namely annotation transfer. Given the operation complexity of labeling an object, how to rationally use the labeled ground truths to reduce the number of operations is a very important issue. Here we mainly use two kinds of annotation transfer, called inter-object transfer and inter-frame transfer.

First, since objects on the same lane, like car, van and cyclist, usually share similar orientation and height, inter-object transfer can significantly reduce the rotation and height adjustment of such boxes, while also making labeling more reasonable in the regions with sparse point clouds.

As for the inter-frame annotation transfer, when labeling consecutive frames, there are usually only slight shifts and deflections between the annotations of these frames, so the operations involved in height adjustment can be greatly reduced by passing labels. Through this kind of transfer, we can avoid the situation of missing labels due to the sparse local point cloud of individual frames as much as possible. Furthermore, we can achieve a one-to-one correspondence between the annotations of consecutive frames, which enables our labels to be used for both 3D object detection and 3D object tracking.

When implementing the transfer, we just copy and paste the labels to minimize the computational overhead of this function. Actually, real-time hand-crafted algorithms hardly avoid other necessary operations like shifting and resizing in the bird view, but usually introduce additional costs. Table~\ref{tab: operations} compares the number of basic operations that will be involved in the case with and without the assistance of annotation transfer, and it can be seen that annotation transfer can significantly reduce the number of operations required in step 3, especially the fine-tuning of height. Therefore, the more consecutive frames a sequence contains, the more objects there are in the same lane in a frame, the more efficient the labeling will be.


\section{Evaluation}
\label{sec:evaluation}

In this section, we will present our evaluation details, including the experimental setup, adopted metrics, and experimental results.

\begin{table}
	\begin{center}
	\begin{tabular}{c|c|c}
	\hline
	Operation & w/o transfer & w/ transfer\\
	\hline\hline
	Locating & $\surd$ & Only shift\\
	\hline
	Rotating & $\surd$ & Little\\
	\hline
	Adjusting box size (BV) & $\surd$ & Little\\
	\hline
	Adjusting height & $\surd$ & Little\\
	\hline
	\end{tabular}
	\end{center}
           \vspace{2ex}
	\caption{Comparison of operations needed with and without annotation transfer. The operations involved when adopting annotation transfer are much easier than otherwise, because we just need to annotate the residuals in most of the time. Note that adjusting box size here refers to adjusting the bird view (BV) projection of the 3D bounding box in the front view and side view of local point clouds.}
	\label{tab: operations}
\end{table}

\begin{table*}
	\begin{center}
	\begin{tabular}{c|c|c|c|c|c}
	\hline
	Multimodal & Inter-object transfer & Inter-frame transfer & Time (s) & BEV IoU (\%) & 3D IoU (\%)\\
	\hline\hline
	$\times$ & $\times$ & $\times$ & 31.10 & 79.30 & 73.85\\
	\hline
	$\surd$ & $\times$ & $\times$ & 25.59 & 85.48 & 81.98\\
	\hline
	$\surd$ & $\surd$ & $\times$ & 21.90 & 86.62 & 83.26\\
	\hline
	$\surd$ &$\surd$ & $\surd$ & \textbf{12.37} & \textbf{87.35} & \textbf{83.73}\\
	\hline
	\end{tabular}
	\end{center}
           \vspace{2ex}
	\caption{Comparison on efficiency and accuracy. Our method finally achieves a 2.5 times speed-up compared with the baseline, and improves the IoU in the bird view and 3D IoU by 8.05\% and 9.88\% respectively. The better improvement on 3D IoU shows that height annotation can benefit a lot from our method.}
	\label{tab: time_IOU}
           \vspace{1ex}
\end{table*}

\begin{table*}
	\begin{center}
	\begin{tabular}{c|c|c|c|c|c|c}
	\hline
	Multimodal & Inter-object transfer & Inter-frame transfer & BEV AP (0.7) & BEV AP (0.5) & 3D AP (0.7) & 3D AP (0.5)\\
	\hline\hline
	$\times$ & $\times$ & $\times$ & 71.43 & 99.62 & 59.74 & 89.48\\
	\hline
	$\surd$ & $\times$ & $\times$ & 88.49 & 99.75 & 86.33 & 90.51\\
	\hline
	$\surd$ & $\surd$ & $\times$ & 87.97 & 99.75 & 86.89 & 89.62\\
	\hline
	$\surd$ &$\surd$ & $\surd$ & \textbf{89.17} & \textbf{99.92} & \textbf{87.24} & \textbf{99.43}\\
	\hline
	\end{tabular}
	\end{center}
           \vspace{2ex}
	\caption{Comparison on average precision. Our method finally increases the AP (0.7) in the bird view and 3D AP (0.7) by 17.74\% and 27.5\% respectively. The improvement on the 3D metrics is also better than that on the metrics computed in the bird view, which further shows the superiority of our method in terms of height annotation.}
	\label{tab: AP}
\end{table*}

\subsection{Experimental Setup}
Although we can intuitively feel that our method improves the efficiency and accuracy of this annotation task, we still tried to test the productivity advances quantitatively and precisely. In each group of experiments, we assigned the randomly selected data from KITTI raw data to the same number of volunteers, and compared the accuracy and efficiency of their annotation. The KITTI dataset provides data from consecutive frames in different scenes, including RGB images, GPS/IMU data, 3D object tracklet labels and calibration data. These data cover six categories of scenes, including city, residential, road, campus, \emph{etc.}, and eight categories of objects. We randomly selected six sequences of different scenes, and five consecutive frames of data from each sequence as our test data. This test benchmark contains a total amount of 374 instances. More detailed analysis of data distribution is shown in Figure~\ref{fig: data_analysis}.

We set up four experimental groups. First of all, we added the function of annotating 3D bounding boxes on top of the open-source tool~\cite{LATTE}, which is a point cloud annotation tool only for the 2D annotation on the bird view. With our supplemented functions, annotators can use this tool to adjust the top and bottom of boxes, and thus we take it as the baseline of our experiments. This method can realize the most basic functions of 3D annotation, but due to the lack of effective organization of multimodal data and full use of data characteristics, it cannot fully realize the complete idea of FLAVA. On this basis, we added various functions of multimodal data, inter-object annotation transfer and inter-frame annotation transfer in turn, as the other three experimental groups, to test the contribution of each function to annotation efficiency and accuracy. The functions of using multimodal data include finding and primarily localizing objects by the RGB image, adjusting and verifying the annotated box by the projected view of local point clouds, and finally verifying annotation results by the RGB image.

We invited the same number of different volunteers to label in each experimental group, to ensure that everyone only used the corresponding features to label, and would not get benefit from improved familiarity and proficiency of annotating these data. Volunteers were asked to only annotate the instances for which they felt confident. For instance, for very distant objects, like cars farther than 70 meters away, because the points that can be obtained from LiDAR are very sparse, they will not be labeled. This reduces the uncertainty of the comparison of results that may be produced due to unreasonable samples. We only verify the instances with corresponding ground truths when evaluating. Specifically, we only evaluate the accuracy of annotated boxes that can intersect with a ground truth.

\subsection{Metrics}
When evaluating the quality of annotation quantitatively, we used different metrics to test the efficiency and accuracy of annotation. For the efficiency of annotation, on the one hand, according to Table~\ref{tab: operations}, we can have a qualitative sense of the operation complexity involved in the annotation process; on the other hand, we used the average time spent when annotating each instance as the metrics to measure the efficiency in practical use.

For the evaluation of accuracy, first of all, we need to note that considering that KITTI’s annotation does not include all instances in a scene, especially the objects behind the drive, we referred to the method of~\cite{LATTE}, asked an expert annotator to provide high-quality annotation as the ground truth of the given test data. We used two metrics to evaluate the accuracy, which are commonly used in 3D object detection: intersection over union (IoU) and average precision (AP). Among them, IoU is only calculated when the object is labeled and has a ground truth at the same time, which is different from average precision. IoU can effectively evaluate the average accuracy of labels that are relatively correct, while average precision can evaluate whether the annotation can identify those objects of interest correctly. When computing average precision, we set two kinds of difficulties. In the relatively strict case, we take the label with IoU greater than 0.7 for car and van while 0.5 for pedestrian and cyclist as a true positive; and the relatively easy standard is that the label with IoU greater than 0.5 for car and van while 0.25 for pedestrian and cyclist can be regarded as a true positive. We also calculated the IoU and AP of 2D boxes in the bird view in addition to 3D boxes, which can help us to analyze the effect of different functions on the most difficult part in this annotation task --- height annotation.

\vspace{1ex}
\subsection{Results}
\textbf{Quantitative Analysis}\quad Since there is no open source tool with similar functions, we supplemented the functions of~\cite{LATTE} so that it can have theoretically complete functions in 3D annotation. We regard it as the baseline of FLAVA. On this basis, we add functions in turn, so that the whole process and functions gradually approach our method. It can be seen from Table~\ref{tab: time_IOU} and \ref{tab: AP} that although it takes the longest time, 31.1s, to annotate each instance in the baseline, its label quality of both 3D and bird-view 2D boxes is poorest under multiple metrics of IoU and average precision.

Subsequently, we firstly organize multimodal data effectively, and we can see that not only the average time used to annotate each instance is reduced by about 6s, but also the IoU and average precision are significantly improved. Moreover, it can be seen that since our height adjustment is mainly implemented in the projected view of the local point cloud, the performance improvement of 3D boxes is much greater than that of 2D boxes in the bird view.

Then we add inter-object transfer and inter-frame transfer, which further improve the accuracy and efficiency of annotation. In particular, introducing inter-frame transfer almost doubles the efficiency of annotation and shows a 2.5 times speed-up compared with the baseline. Note that this improvement is achieved on our specific test benchmark, where a sequence only consists of 5 consecutive frames. It is conceivable that the more frames a sequence contains, the greater this improvement will be. Furthermore, annotation transfer also makes the height annotation more stable and accurate. It can be seen that 99.75\% of AP(0.5) in the bird view of the 2nd group of experiments is not much different from 99.92\% of the 4th group, but 90.51\% is much lower than 99.43\% in terms of the 3D AP. Similar improvements brought by annotation transfer can also be reflected in other metrics results. Finally, compared to other public annotation tools, the accuracy outperforms \cite{3DBAT} (about 20\% 3D IoU) by a large margin and the user experience is considered to be smoother from all of our volunteers' feedback.

\begin{figure*}
   \centering
   \includegraphics[width=0.87\linewidth]{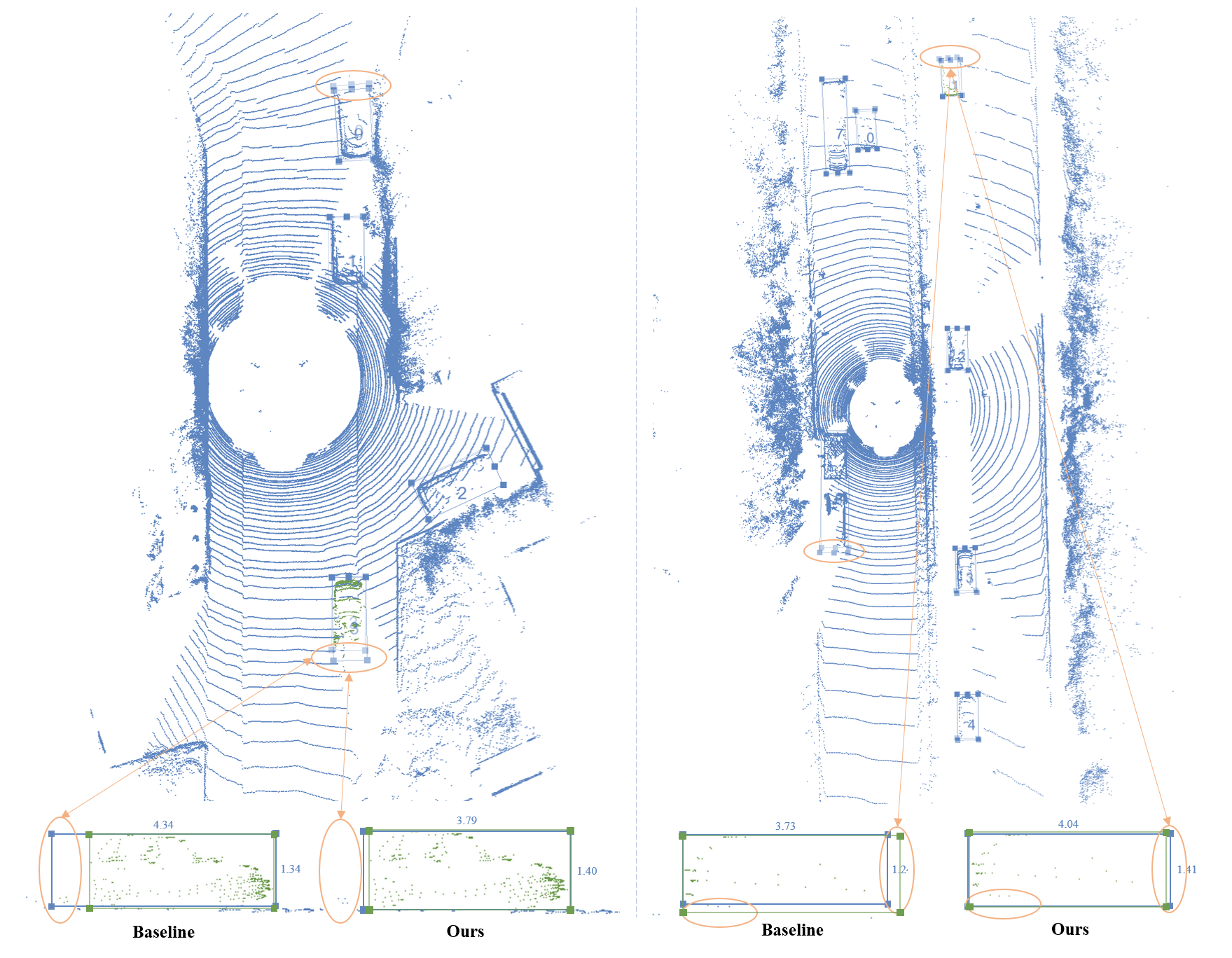}
   \caption{Qualitative analysis of annotation results. The two examples show the different annotations from the baseline and our method, where the annotation from the baseline is plotted as a more transparent background in the figures to show their difference. We select two annotated objects for comparison. Annotated boxes and ground truths are marked in blue and green respectively. Points inside the annotated box are highlighted in green. It can be seen that the error often happens when annotating the front, back and bottom of the object.}
   \label{fig: qualitative}
\end{figure*}

\textbf{Qualitative Analysis}\quad To have a more intuitive understanding of the improved label quality, we show some examples to compare the annotations from the baseline and our proposed method (Figure~\ref{fig: qualitative}). Firstly, it can be seen that from the bird view, there exist some slight but noticeable differences when annotating the front and the back of cars. In the left example, there are some noises behind the car, which are not clear from the bird view. However, our adjustment in the side view can help a lot. Similarly, the bottom of the car in the right example adjusted from the side view is more accurate than that adjusted from the perspective view. Furthermore, due to the annotation transfer adopted in our method, the front of the car is consistent with the more confident annotation in previous frames, which is also more accurate.

\begin{figure*}
   \centering
   \includegraphics[width=0.87\linewidth]{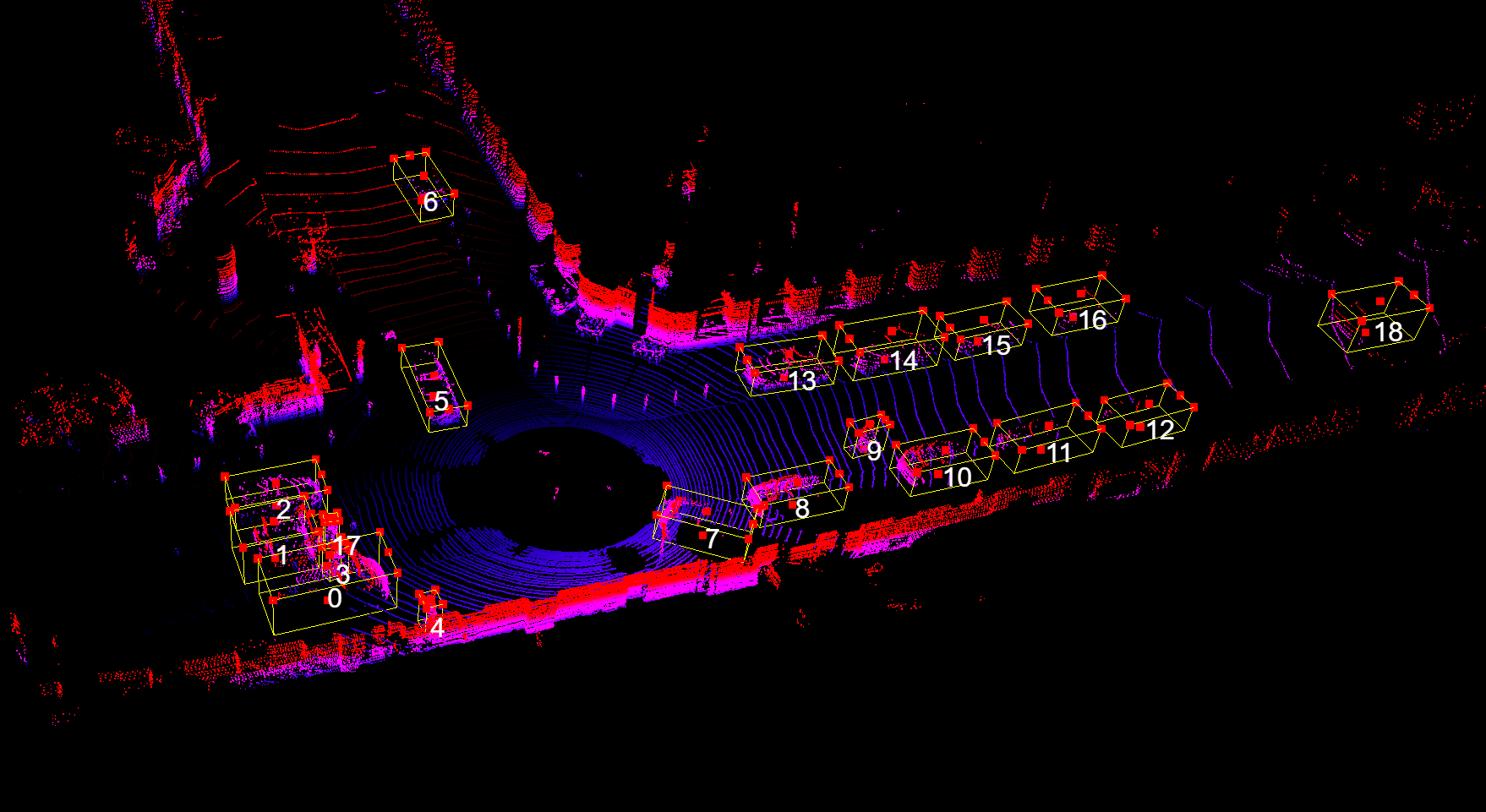}
   \caption{An example of our annotation results. Best viewed in color.}
   \label{fig: result_example}
\end{figure*}

In a word, from both quantitative and qualitative results, it can be seen that the performance of baseline based on 3D interaction can be greatly improved by leveraging the multimodal data due to its contribution to the better identification of distant objects and the more accurate annotation of box boundaries.
The introduction of annotation transfer fully utilizes the specific characteristics of data. It further improves the efficiency and accuracy of annotation, making the whole annotation procedure more constructive and flexible. An example of our annotation results is shown in Figure~\ref{fig: result_example}. See more examples of our annotation process and results in the demo video.

\section{Discussion}
From the previous discussion and evaluation, it is evident that all the annotated operations and verification should not be performed only on a single modal data. We need to give full consideration to what kind of data is more appropriate for the operation of annotator and what kind of data can highlight the saliency of our interest. A constructive pipeline and the effective organization of multimodal data can greatly improve the efficiency and accuracy of annotation. At the same time, novel algorithms are sometimes not very practical regarding the equipment given to the annotators in application. Instead, combining various simple but efficient techniques may be more effective in improving the user experience of annotators.

Although our FLAVA solves some basic problems in labeling point cloud, there are still challenges in application. First, labeling point cloud is a relatively skilled work. In the actual annotation process, many annotators have received professional training and long-term practice to further improve their efficiency and proficiency. Therefore, it is interesting if there is a way to use our annotation tools to train them pertinently. Maybe it can achieve unexpected results while reducing the training workload. Similarly, we can also use active learning to improve the performance of related algorithms efficiently through the interactions between annotators and tools. These are some possibilities that can be mined in this interaction procedure.

In addition, there are some other engineering problems in application. For example, when the number of points becomes larger, whether it will affect the performance of our annotation tool. The test result is that for the current web-based tool, about 100 thousand of point cloud data can be imported quickly enough. About 1000 thousand of point cloud data takes nearly half a minute to import without affecting the interactive process of annotation. When the resolution of the input point cloud becomes further higher, the time of importing data and the fluency of operation may also become important factors restricting the tool. Another engineering problem is the synchronization of different modal data. Sometimes the image and point cloud data cannot be fully synchronized. How to solve the impact of this deviation on the annotation process is also worth further exploration. Finally, although we propose a systematic annotation process for the task of 3D object detection and tracking, there still exist new difficulties in other annotation tasks like point cloud semantic segmentation, which may also need specific designs tailored to those tasks.

In the process of annotation, we also try to get the inspiration for the current 3D detection algorithms. For example, human beings usually verify the annotation results in RGB images, which has not been well modeled and applied in the detection algorithms. On the other hand, human annotation quality may be regarded as an important goal and performance bottleneck of LiDAR-based object detection algorithms. The current state-of-the-art methods can achieve about 80\% of 3D AP (0.7) without considering the efficiency of the algorithm when detecting cars, while our annotation can achieve about 90\%. Therefore, the gap between current algorithms and human's ability can be estimated roughly. How to further reduce this gap is a problem that researchers need to consider at present. At the same time, when the gap is closed, it may also indicate that the point cloud data has been utilized to the greatest extent, and further considering the combination with other data and control algorithms may be a more important task.

\section{Conclusion}
In this paper, we propose FLAVA, a systematic annotation method to minimize human interaction when annotating LiDAR-based point clouds. It aims at helping annotators solve two key problems, identifying the objects of interest correctly and annotating them accurately.
We carefully design a UI tailored to this pipeline and introduce annotation transfer regarding the specific characteristics of data and tasks, which make annotators be able to focus on simpler tasks at each stage and accomplish it with fewer interactions.
Detailed ablation studies demonstrate that this annotation approach can effectively reduce unnecessary repeated operations, and significantly improve the efficiency and quality of annotation. At last, we discuss the various thinking and possibilities of the extension of this annotation task. Future work includes designing annotation tools for other tasks upon LiDAR-based point clouds and improving related algorithms based on human’s annotation procedure.

\label{sec:conclusion}

\bibliographystyle{SIGCHI-Reference-Format}
\bibliography{sample}

\end{document}